\documentclass[twoside]{article}

\usepackage[width=21cm,height=29.7cm,left=2cm,right=2cm,top=2cm,bottom=2cm]{geometry}

\usepackage[round]{natbib}

\usepackage {tikz}  % for tikz figures
\usetikzlibrary {positioning}
\usetikzlibrary{fit}
\usetikzlibrary{backgrounds}
\usepackage{amssymb}% for \mathbb (board bold)
\usepackage{enumitem}
\usepackage{amsmath}
\usepackage{xspace}

\definecolor{mygreen}{rgb}{0.2, 0.7, 0.2}
\definecolor{myorange}{rgb}{0.9, 0.5, 0.0}

\definecolor{blue}{HTML}{5DA5DA}
\definecolor{orange}{HTML}{FAA43A} 
\definecolor{green }{HTML}{60BD68} 
\definecolor{pink  }{HTML}{F17CB0} 
\definecolor{brown }{HTML}{B2912F} 
\definecolor{purple}{HTML}{B276B2} 
\definecolor{yellow}{HTML}{DECF3F} 
\definecolor{red   }{HTML}{F15854} 
\definecolor{gray  }{HTML}{4D4D4D}

\newcommand{\ff}{\boldsymbol{f}}
\renewcommand{\gg}{\boldsymbol{g}}

\renewcommand{\tt}{\boldsymbol{t}}
\newcommand{\xx}{\boldsymbol{x}}
\newcommand{\yy}{\boldsymbol{y}}
\newcommand{\zz}{\boldsymbol{z}}

\newcommand{\XX}{\boldsymbol{X}}
\newcommand{\YY}{\boldsymbol{Y}}
\newcommand{\ZZ}{\boldsymbol{Z}}

\newcommand{\ddelta}{\boldsymbol{\delta}}
\newcommand{\vvarepsilon}{\boldsymbol{\varepsilon}}

\newcommand{\eeta}{\boldsymbol{\eta}}

\newcommand{\ttheta}{{\boldsymbol{\theta}}}

\newcommand{\Pphi}{\boldsymbol{\Phi}}
\newcommand{\zzero}{\boldsymbol{0}}

\newcommand{\Dcal}{\mathcal{D}}
\newcommand{\Ecal}{\mathcal{E}}
\newcommand{\Lcal}{\mathcal{L}}

\newcommand{\Ncal}{\mathcal{N}}

\newcommand{\Rset}{\mathbb{R}}
\newcommand{\esp}{\mathbb{E}}
\newcommand{\cov}{\mathrm{cov}}
\newcommand{\var}{\mathrm{var}}
\newcommand \trans {^\top}
\newcommand \drm {\mathrm{d}}

\newcommand{\E}{\mathrm{E}}

\newcommand{\norm}{\mathcal{N}}

\newcommand{\wvect}{\mathbf{w}}

\newcommand{\bigO}{\mathcal{O}}

\newcommand{\name}[1]{{\textsc{#1}}\xspace}

\newcommand{\mcmc}{\name{mcmc}}

\newcommand{\koh}{\textsc{koh}\xspace}

\newcommand{\lagp}{\name{lagp}}

\newcommand{\gp}{\name{gp}}
\newcommand{\gps}{\textsc{gp}s\xspace}
\newcommand{\dgp}{\name{dgp}}
\newcommand{\dgps}{\textsc{dgp}s\xspace}

\newcommand{\mse}{\name{mse}}

\newcommand{\vcal}{\name{v-cal}}

\newcommand{\svi}{\name{svi}}

\definecolor {processblue}{cmyk}{0.2,.1,0.1,.0}
\definecolor {randomblue}{cmyk}{0.57,0.05,0,0.76}
\definecolor {lightleafyellow}{cmyk}{.00,.29,.67,.68}
\definecolor {leafyellow}{cmyk}{.00,.30,.98,.30}
\definecolor {treeorange}{cmyk}{.00,.67,.83,.09}
\definecolor {grassgreen}{cmyk}{.99,.0,.23,.40}
\newdimen\microwidth 
\newdimen\microheight
\newdimen\circlediameter
\newdimen\microwidthB 
\newdimen\microheightB
\newdimen\circlediameterB
\title{Variational Calibration of Computer Models }

\author{
S\'ebastien Marmin\footnote{Department of Data Science, EURECOM, France} \hspace{2cm} Maurizio Filippone\footnotemark[1]
}

\begin{document}

\maketitle

\begin{abstract}
Bayesian calibration of black-box
 computer models offers an established framework to obtain a posterior distribution over model parameters.
Traditional Bayesian calibration involves the emulation of the %
 computer model and an additive model discrepancy term using Gaussian processes; inference is then carried out using MCMC.
These choices pose computational and statistical challenges and limitations, which we overcome by proposing the use of approximate Deep Gaussian processes and variational inference techniques.
The result is a practical and scalable framework for calibration, which obtains competitive performance compared to the state-of-the-art.
\end{abstract}

\section{INTRODUCTION}

The inference of parameters of expensive computer models from data is a classical problem in Statistics \citep{Sacks89}. 
Such a problem is often referred to as {\em calibration} \citep{kennedy2001Bayesian}, and the results of calibration are often of interest to draw conclusions on parameters that may have a direct interpretation of real physical quantities.
There are many fundamental difficulties in calibrating expensive computer models, which we can distinguish between computational and statistical. 
Computational issues arise from the fact that traditional optimization and inference techniques require running the expensive computer model many times for different values of the parameters, which might be unfeasible within a given computational budget. 
Statistical limitations, instead, arise from the fact that computer models are abstractions of real processes, which might be inaccurate.

Building on previous work from \citet{Sacks89}, in their seminal paper, \citet{kennedy2001Bayesian} propose a statistical model, based on Gaussian processes (\gps; \citet{Rasmussen06}), which jointly tackles the problems above.
In their model, which we will refer to as \koh, the output of a deterministic computer model is emulated through a \gp estimated from a set of computer experiments; in this way, computational issues are bypassed by using the prediction of the \gp for a given set of parameters in place of the computer model.  
Observations from the real process, also known as field observations, instead, are modeled as the sum of the output of the \gp emulating the computer model and another \gp, which accounts for the mismatch between the computer model and the real process of interest.
The \koh model is treated in a Bayesian way, making it suitable for problems where quantification of uncertainty is an important requirement. 
This is often the case when one is interested in drawing conclusions on parameters of interest, making predictions for decision-making with specific cost associated to predictions, or when one is interested in iteratively improve the experimental design.

While the \koh model and inference make for an attractive and elegant framework to tackle quantification of uncertainty for calibration of expensive computer models, there are a number of limitations, which we aim to overcome with this work.
From the modeling perspective, \gps are indeed flexible emulators, provided that a suitable covariance function is chosen. 
This may limit the class of functions that can be modeled compared to alternatives from, for instance, the literature of nonstationary \gps (e.g. \citet{paciorek2003nonstationary}) or, more recently, of Deep \gps (\dgps; \citet{Damianou13}). %
From the computational perspective, limitations are inherited from the scalability issues of \gps \citep{Rasmussen06}, which form the building blocks of this model, and the use of Markov chain Monte Carlo (\mcmc) \citep{Neal93} techniques to carry out inference. 

This work aims to tackle these issues by proposing the use of recent developments in the \gp and \dgp literature and variational inference, to 
(i) extend the modeling capabilities of \gps in the emulation using \dgps;
(ii) extend the original framework in \citet{kennedy2001Bayesian}, by casting the model as a special case of a \dgp;
(iii) use techniques based on random feature expansions and stochastic variational inference, building on the work by \citet{Cutajar17}, to obtain a scalable framework for Bayesian calibration of computer models.
We extensively validate our proposal, which we name \vcal, on a variety of calibration problems, comparing with alternatives from the state-of-the-art. 
We demonstrate the flexibility and the scalability of \vcal, as well as the ability to capture the uncertainty in model parameters and model mismatch. 

\section{RELATED WORK}

The framework of \citet{kennedy2001Bayesian} and related calibration methodologies (e.g., \citet{bayarri2007Framework,higdon2004combining,goldstein2004Probabilistic}) have been extended and developed for applications in fields as diverse as climatology \citep{sanso2008Inferring,salter2018Uncertainty}, environmental sciences \citep{larssen2006forecasting,arhonditsis2007Eutrophication}, biology \citep{henderson2009bayesian}, and mechanical engineering \citep{williams2006Combining}, to name a few. %

More generally, Bayesian calibration had a large echo in statistical analysis of computer experiments.
Among many others, 
\citet{wang2009Bayesian} propose a Bayesian approach where the posterior of the field observation is expressed as the sum of two independent posterior distributions derived separately, representing the discrepancy and the computer model. % 
\citet{storlie2015Calibration} propose a calibration approach which can handle categorical inputs within a Bayesian smoothing spline ANOVA framework.

Identifiability issues with the \koh model were pointed out in the discussion of the paper of \citet{kennedy2001Bayesian}. 
Such issues arise due to the over-parameterization of the model, whereby it is possible to confound the effects of calibration parameters and model discrepancy.
Later, more work has been done on illustrating the lack of identifiability and proposing ways to mitigate it \citep{arendt2012improving,brynjarsdottir2014learning}. 
In the former, %
it is shown that, when the computer model has multiple responses, identifiability is improved by considering them jointly with multiple output \gp models.
In the latter, the authors show with simple examples the crucial importance of prior information, and they discuss how eliciting this may be difficult for arbitrary black-box computer models.

\citet{tuo2016Theoretical} propose a proof of inconsistency of a simplified version of the \koh model for Bayesian calibration. 
As an alternative, calibration with convergence properties is performed by minimizing a loss function involving the discrepancy.
More recently, \citet{plumlee2017bayesian} provide analytic guarantees on the use of discrepancy priors that reduce the suboptimal broadness of the posterior. %

A large literature is devoted to the practicalities of numerically challenging applications.    
The \koh calibration model and its inference suffer from the well known computational burden of standard \gp models. 
\citet{higdon2008computer} use basis functions representations to tackle problems with high dimensional output.
\citet{gramacy2015Calibrating} use local approximate \gp modeling and calibrate parameters by solving a derivative-free maximization of a likelihood term.
\citet{pratola2016Bayesian} handle large problems using a sum-of-trees regression for modeling jointly data from computer model and the real process.

Very recent approaches combine Bayesian and discrepancy-based loss minimization providing good performance on moderate-size data sets. 
\citet{gu2018Scaled} perform calibration within a Bayesian framework, %
by defining a prior distribution directly on the $L_2$ norm of the discrepancy. % 
Finally, \citet{wong2017frequentist,xie2018Bayesian} %
sample from the posterior distribution over calibration parameters by minimizing the $L_2$ norm of a sample path of the discrepancy. %

\section{VARIATIONAL CALIBRATION}

In this section we introduce \vcal, which is based on the \koh calibration model, random feature expansions of \gps and \dgps, and Stochastic Variational Inference (\svi). %
We first review the \koh model for Bayesian calibration of computer models and random feature expansions of \gps and \dgps. 
We then present our contribution; see the supplementary material for an illustrative example describing how \vcal works in detail. 

\subsection{Background on Bayesian Calibration}
Consider prediction and uncertainty analysis of a physical phenomenon approximated by a computer model (often expensive to evaluate).
Observations $Y= \left[\yy_1,\ldots, \yy_n\right]\trans$ $\in \Rset^{n\times d_\text{out}}$ are made over variable inputs $X=[\xx_1,\ldots,\xx_n]\trans$ in a given set $\Dcal_1\subset\Rset^{d_1}$. 
For example, in a climatological context, $Y$ could correspond to temperature measurements with respect to the latitude and the longitude (with $d_\text{out}=1$ and $\Dcal_1 = [-90,90]\times [-180,180[$).
The computer model simulating the real phenomenon requires the calibration inputs $\ttheta\in\Dcal_2\subset\Rset^{d_2}$, and, given these, it is a function of $\xx\in \Dcal_1$. 
Calibration inputs $\ttheta$ determine which specific application we are reproducing (e.g., exchange rates determining the carbon cycle). 
We use $\tt \in \Dcal_2$ to denote particular values of calibration inputs, and $\ttheta$ to denote the true (unknown) values we are interested in inferring.
In their Bayesian formulation, \citet{kennedy2001Bayesian} introduce a prior over $\ttheta$, and they aim to characterize the posterior distribution over $\ttheta$ given the data collected from the observations of the real process and runs of the computer model.

Beside the observations $Y$ associated with $X$, the computer model is run at (possibly different) inputs $X^* = \left[\xx_1^*,\ldots,\xx_N^*\right]\trans$ with calibration inputs $T=\left[\tt_1,\ldots,\tt_N\right]\trans$. 
Generally $N$ is larger than $n$ as running the computer model is easier (albeit computationally expensive) compared to performing a real world observation. 
We denote by $Z=\left[\zz_1,\ldots,\zz_N\right]\trans\in \Rset^{N\times d_\text{out}}$ the output of computer model evaluated at $X^*$ and $T$.

\begin{figure}[t]
\vspace{.0in}
\begin{center}
\begin {tikzpicture}[-latex ,auto ,node distance =\microwidthB and \microheightB ,on grid ,
semithick ,
stateData/.style ={ circle ,top color =white , bottom color = leafyellow!20 ,draw,leafyellow , text=leafyellow , minimum width =\circlediameterB},
stateNoth/.style ={},
statePhi/.style ={ circle ,top color =white , bottom color = lightleafyellow!20 ,draw,lightleafyellow , text=lightleafyellow , minimum width =\circlediameterB},
stateNoth/.style ={},
stateThe/.style ={ circle ,top color =white , bottom color = treeorange!20 ,draw,treeorange , text=treeorange , minimum width =\circlediameterB},
stateNoth/.style ={},
stateFun/.style ={ circle ,top color =white , bottom color = grassgreen!20 ,draw,grassgreen , text=grassgreen , minimum width =\circlediameterB},
stateNoth/.style ={},
stateW/.style ={ rectangle ,top color =white , bottom color = grassgreen!20 ,draw,grassgreen , text=grassgreen , minimum width =1.5\circlediameterB},
stateNoth/.style ={},
statePathPts/.style ={inner sep = -10cm},
myrect/.style={rectangle, draw=grassgreen!20, inner sep=0pt,line width=2pt,fit=#1}]
\node[stateNoth] (C)
{};
\node[stateData] (A1) [above left=of C] {$\xx$};
\node[stateNoth] (Noth1) [below right=of A1] {};
\node[stateData] (A1) [above left=of C] {$\xx$};
\node[stateNoth] (Noth9) [below right=of Noth1] {};
\node[stateThe] (Theta) [below right=of Noth1] {$\ttheta$};
\node[stateNoth] (Noth11) [right=of Theta] {};
\node[stateNoth] (Noth12) [right=of Noth11] {};
\node[statePathPts] (Noth13) [above=of Noth12] {};
\node[stateNoth] (Noth2) [below left=of Theta] {};
\node[stateData] (A2) [below left=of Noth2] {$\xx$};
\node[stateNoth] (Noth3) [below=of A2] {};
\node[stateNoth] (Text1) [below left=of Noth2] {\color{leafyellow}$~~~{}^*$};
\node[stateData] (A3) [below=of Noth3] {$\tt$};
\node[stateNoth] (Noth4) [right=of A1] {};
\node[stateNoth] (Noth5) [right=of Noth4] {};
\node[stateNoth] (Noth6) [right=of Noth5] {};
\node[stateNoth] (Noth7) [right=of Noth6] {};
\node[stateNoth] (Noth8) [right=of Noth7] {};
\node[stateNoth] (Noth10) [right=of Noth8] {};
\node[stateW] (Delta) [above=of Noth10] {$\ddelta(\cdot)$};
\node[stateW] (Eta) [below=of Noth10] {$\eeta(\cdot,\cdot)$};

\node[stateNoth] (C1brtem) [right=of Eta] {};
\node[stateNoth] (C1br) [below right=of C1brtem] {};
\node[stateNoth] (C1altem) [left=of Delta] {};
\node[stateNoth] (C1al) [above left=of C1altem] {};
\node[myrect={(C1al) (C1br)}] (RECT1) {}; 

\node[stateNoth] (N1) [right=of A3] {};
\node[stateNoth] (N2) [right=of N1] {};
\node[stateNoth] (N3) [below=of N1] {};
\node[stateNoth] (N4) [right=of N2] {};

\draw (Theta.east) -- (RECT1.221);%(1.5,-0.5) |- (Eta.195);
\draw[densely dotted,-] (RECT1.221) -|  (1.6,-0.15) ;
\draw[densely dotted]   (1.6,-0.15) --  (Eta.195);
\draw (A1.east) -- (RECT1.west);%
\draw[densely dotted] (RECT1.west) -| (1.6,0.45) |- (Eta.165);
\draw[densely dotted] (1.6,0.45) |-  (Delta.west);

\draw[densely dotted, -] (Delta.east) -|  (3.4,0.45);
\draw[densely dotted, -] (Eta.east) -|  (3.4,0.45);
\draw[densely dotted, -] (3.4,0.45) |-  (RECT1.0);

\node[stateNoth] (RECT1EAST) [above right=of C1brtem] {};
\node[stateNoth] (N001) [right=of RECT1EAST] {};
\node[stateNoth] (N002) [right=of N001] {};
\node[stateData] (B1) [right=of N002] {$\yy$};

\node[left] at (RECT1EAST) {\tiny$+$};

\draw (RECT1.0) --  (B1);

\node[stateNoth] (NN01) [below right=of A2] {};
\node[stateNoth] (NN02) [right=of NN01] {};
\node[stateNoth] (NN03) [right=of NN02] {};
\node[stateNoth] (NN04) [right=of NN03] {};
\node[stateNoth] (NN05) [right=of NN04] {};
\node[stateNoth] (Eta2) [right=of NN05] {};
\node[stateNoth] (C2br) [below right=of Eta2] {};
\node[stateNoth] (C2al) [above left=of Eta2] {};
\node[myrect={(C2al) (C2br)}] (RECT2) {}; 

\node[stateNoth] (RECT2EAST) [right=of Eta2] {};
\node[stateNoth] (N001z) [right=of RECT2EAST] {};
\node[stateNoth] (N002z) [right=of N001z] {};
\node[stateNoth] (N003z) [right=of N002z] {};
\node[stateData] (B1z) [right=of N003z] {$\zz$};

\draw (RECT2.0) --  (B1z);

\draw[-] (A2) -|  (0.75,-1.85);
\draw[-] (A3) -|  (0.75,-2.4);
\draw (0.75,-1.85) |-  (RECT2.170);
\draw (0.75,-2.4) |-  (RECT2.190);
\draw[-,densely dotted] (RECT2.170) -|  (Eta2);
\draw[-,densely dotted,on background layer] (RECT2.190) -|  (Eta2);
\draw[-,densely dotted,on background layer] (Eta2.0) -- (RECT2.0);
\node[stateW] (Eta2real) [right=of NN05] {$\eeta(\cdot,\cdot)$};
\node[stateNoth] (Noo2) [above=of Eta2] {};
\node[above] at (Noo2) {\large$\substack{\text{Computer code} \\~}$};
\node[stateNoth] (Noo) [above=of Delta] {};
\node[above] at (Noo) {\large$\substack{\text{Natural phenomenon}\\{\color{grassgreen}\ff(\cdot,\cdot)} \\~}$};
\node[above] at (Delta) {\large$\substack{\text{Discrepancy}\\~}$};
\end{tikzpicture}
\vspace{-5mm}
\end{center}
\vspace{.0in}
\caption{\koh Calibration Model.}
\label{fig:concept}
\end{figure}
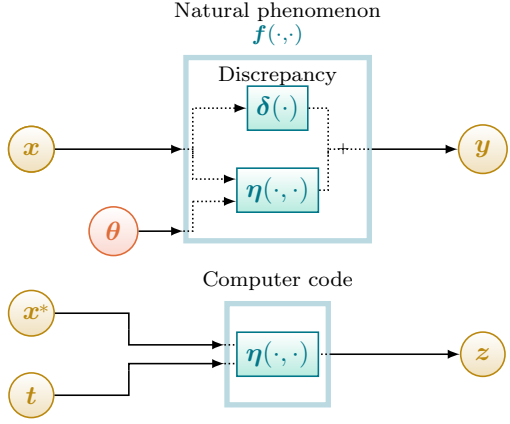

We assume that $Y$ and $Z$ are drawn from some distributions $p(\yy_i|\ff_i)$ and $p(\zz_j|\eeta_j^*)$, which determine the likelihood functions. 
The matrices $F=[\ff_1,\ldots,\ff_n]\trans$ and $H^*=[\eeta_1^*,\ldots,\eeta_N^*]\trans$ result from mapping $(\xx_i,\ttheta)_{i=1,\ldots,n}$ and $(\xx_j^*,\tt_j)_{j=1,\ldots,N}$ through $\ff$ and $\eeta$, respectively. %
The link between the computer model (with latent representation $\eeta$) and the real phenomenon (with latent representation $\ff$) is modeled by
\begin{equation}
%% \forall (\xx, \tt) \in \Dcal_1\times\Dcal_2,~~
\ff(\xx,\tt) = \eeta(\xx,\tt)+\ddelta(\xx),
\label{eq:model}
\end{equation}
where $\ddelta$ represents the discrepancy between the computer model and the real process. 
Figure~\ref{fig:concept} illustrates the \koh calibration model.

In this framework, $\eeta$ and $\ddelta$ are expressed as independent multidimensional \gps.
In order to keep the notation uncluttered, we denote the \gp covariance parameters for $\eeta$ and $\ddelta$ with $\psi$ and the input locations $X$, $X^*$, $T$ with $\Xi$.
The marginal likelihood is %
\begin{align}
p(\left.Y,Z \right|\Xi,\psi) &= \int_{}
p(\left.Y\right|H_{\ttheta}\!+\!\Delta)p(\left.Z\right|H^*)p(\left.\Delta\right|X,\psi)\nonumber\\&p(\left.H_{\ttheta},H^*\right|\ttheta,\Xi,\psi)p(\ttheta)
\drm H^* \, \drm \Delta \, \drm H_\ttheta \, \drm \ttheta\nonumber,
\end{align}
where % 
$H_{\ttheta}=[\eeta(\xx_1,\ttheta),\ldots,\eeta(\xx_n,\ttheta)]\trans$ and $\Delta = [\ddelta(\xx_1),\ldots,\ddelta(\xx_n)]\trans$.
The nontrivial dependence of this integrand with respect to $\ttheta$, which are the parameters of interest, makes inference over $\ttheta$ intractable, and this requires approximations.
We do this by employing an approximation of the \gps composing the model via random feature expansions, building on the work by \citet{Gal16,Cutajar17}, and through variational inference techniques.

\subsection{Generalized KOH Calibration Model}
The original formulation of the \koh calibration model involves the use of \gps to emulate the computer model and to model the additive discrepancy.
As pointed out by \cite{kennedy2001Bayesian}, additive discrepancy is very specific and this formulation can be relaxed (as e.g. in \citet{qian2008Bayesian}).  
We propose to do so by assuming that observations from the real process are obtained through the warping of the emulator, as follows
\begin{equation}
%% \forall (\xx, \tt) \in \Dcal_1\times\Dcal_2,~~
\ff(\xx,\tt) = \gg\left(\eeta(\xx,\tt),\xx\right).
\label{eq:generalModel}
\end{equation}
Clearly, we retrieve the \koh formulation when % 
the warping applies the identity to $\eeta(\xx,\tt)$ and adds it to a \gp on $\xx$. 
We can therefore interpret the \koh model as a special case of a \dgp, where the first layer implements the emulation of the computer model and the second layer implements a particular type of warping to model the real process. 
Similarly to the original \koh model, its generalization allows one to reason about the mismatch between the computer model and the real process through the analysis of the warping function. 
We will illustrate examples in the experiments.

\subsection{Random Features Expansions}
In order to develop a practical and scalable calibration framework, we propose to approximate \gps using random feature expansions \citep{Rahimi08,LazaroGredilla10,Cutajar17}. 
This approximation turns \gps into Bayesian linear models with a set of basis functions determined by the choice of the \gp covariance function, as discussed next.

Consider a \gp prior with zero mean (without loss of generality) and a covariance function $k(\cdot, \cdot)$. 
The properties of the covariance function determine the characteristics of the \gp prior. 
When the covariance function is Gaussian, Mat\'ern or arc-cosine, to name a few, it is possible to show that draws from the \gp prior are a linear combination of an infinite number of basis functions, with Gaussian-distributed weights \citep{Neal96,Rasmussen06}.
In practice, denoting by $\ff$ a draw from a \gp evaluated at $n$ inputs, this can be expressed as $\ff = \Phi \wvect$, with $\wvect \sim \norm(0,I)$ and infinite dimensional, and $\Phi$ the evaluations of the infinite basis functions at the $n$ inputs. 
The covariance of $\ff$ is readily obtained as 
$$
\cov(\ff) = \E(\Phi \wvect \wvect^\top \Phi^\top) = \Phi \E(\wvect \wvect^\top) \Phi^\top = \Phi \Phi^\top.
$$

The idea of random feature expansions is to obtain tractable ways to truncate the infinite representation of \gps, introducing a finite set of basis functions and a finite set of weights for approximating efficiently $\ff$.
There are various ways to carry out the truncation; when the covariance is shift-invariant,
it is possible to express the covariance as the Fourier transform of a positive measure, and this makes apparent the low-rank decomposition of the covariance considering a finite set of frequencies \citep{Rahimi08}.

When using \gps in modeling problems with $n$ observations, the truncation has the advantage of avoiding the need to solve algebraic operations with the covariance matrix, which generally cost $\bigO(n^3)$ operations.
Instead, the truncation turns \gps into Bayesian linear models, for which inference can be done linearly in $n$ and $\bigO(N_{\mathrm{RF}}^3)$, where $N_{\mathrm{RF}}$ denotes the number of random features used in the truncation. 

\begin{figure}[t]
\vspace{.0in}
\begin {center}
\input{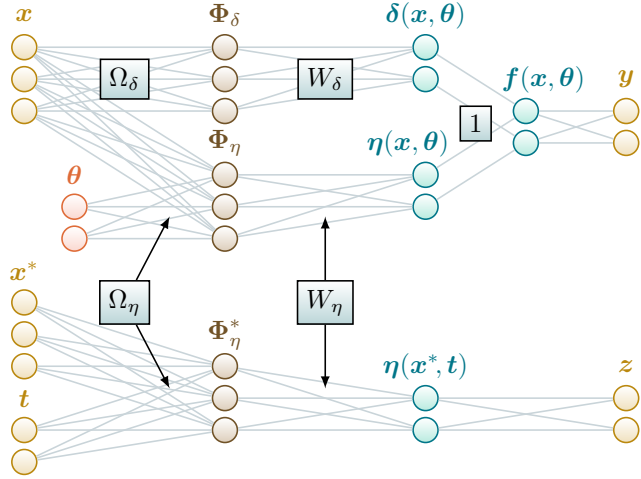}
\end{center}
\vspace{.0in}
\caption{Neural Network representation of the proposed approximation to the \koh model. 
Equations \eqref{eq:rfEta} and \eqref{eq:rfDelta} formulate two-layer modules for \gp approximations of $\eeta$ and $\ddelta$.}
\label{fig:nn}
\end{figure}

In this work, we propose to expand the two \gps $\eeta$ and $\ddelta$ in the \koh calibration model using $N_\text{RF}$ random features. 
Assuming that the \gps have a Gaussian covariance with precision $A_\eta$ and $A_\delta$ and marginal variances $\sigma_\eta^2$ and $\sigma_\delta^2$, 
we obtain:
\begin{align}
\eeta(\xx,\ttheta) & = \Pphi_\eta(\Omega_\eta^{(1)}\xx+\Omega_\eta^{(2)}\ttheta)\trans W_\eta,
\label{eq:rfEta}\\
\ddelta(\xx) & = \Pphi_\delta(\Omega_\delta\xx)\trans W_\delta.
\label{eq:rfDelta}
\end{align}
Here, the functions $\Pphi_\eta,\Pphi_\delta:\Rset^{N_\text{RF}}\rightarrow \Rset^{N_\text{RF}}$ consist in the element-wise application of sine and cosine functions, scaled by $\frac{\sigma_\eta}{\sqrt{N_\text{RF}/2}}$ and $\frac{\sigma_\delta}{\sqrt{N_\text{RF}/2}}$, respectively. % $\phi: \Rset\rightarrow \Rset$. 
The elements of $W_\eta$ and $W_\delta$, of size $N_\text{RF}\times d_\text{out}$, have i.i.d. standard normal priors, whereas the frequency matrices $\Omega_\eta = [\Omega_\eta^{(1)},\Omega_\eta^{(2)}],$ $\Omega_\delta,$ of size $N_\text{RF}\times (d_1+d_2)$, $N_\text{RF}\times d_1$, have i.i.d. normal rows, with covariance dependent on the  positive definite matrices $A_{\eta}$ and $A_{\delta}$; %
in particular, each row $\Omega_\eta$ is i.i.d. $\Ncal(\zzero,A_\eta)$.
Similar considerations apply to $\Omega_{\delta}$. %
Figure~\ref{fig:nn} represents the model (using a neural network-like diagram) according to Equations~\eqref{eq:model}, \eqref{eq:rfEta}, and~\eqref{eq:rfDelta}.

\paragraph{Deep extension:}
It is possible to extend the proposed formulation letting $\eeta$ and/or $\ddelta$ to be modeled as \dgps instead of \gps. 
This is straightforward, as the random feature approximation turns \gps into shallow Bayesian neural networks, so it is possible to approximate \dgps by stacking these approximate \gps, obtaining a Bayesian deep neural network. 
The deep extension is particularly useful when the emulator or the real process exhibit space-dependent behavior that are difficult to model by designing appropriate covariance functions. 
\dgps offer a way to learn such nonstationarities from data, so this is particularly appealing in such challenging applications. 
We will explore this possibility in the experiments.

\subsection{Stochastic Variational Inference}
In this work, we propose a general formulation based on variational inference techniques to approximate the posterior distribution over all model parameters, that is $W_{\eta}$, $W_{\delta}$, and $\ttheta$, noting that there might be cases where $W_{\eta}$, $W_{\delta}$ can be inferred analytically (Gaussian likelihoods).
In particular, we introduce an approximation to the posterior $q(W_\eta,W_\delta,\ttheta)$, which we aim to make as close as possible to the actual posterior over these parameters. 
Using standard variational inference techniques, it is possible to derive a lower bound of the log-marginal likelihood as follows:
\begin{equation}
\Lcal %\log \left(p(Y,Z|\Xi,\psi,\Omega)\right)\\
\geq \Ecal - \mathrm{D}_\text{KL}(q(W_\eta,W_\delta,\ttheta)||p(W_\eta)p(W_\delta)p(\ttheta)) \text{,}
\label{eq:lowerBound}
\end{equation}
where
$
\Ecal\! =\! \esp_{q(W_\eta,W_\delta,\ttheta)}\!\left[\log (p(Y,Z|\Xi,\psi,\Omega,W_\eta,W_\delta,\ttheta))\right]
$ and $\Omega=[\Omega_\eta,\Omega_\delta]$.
With an expression for the lower bound of the marginal likelihood, we can now attempt to maximize it with respect to the parameters of $q(W_\eta,W_\delta,\ttheta)$. 
The lower bound contains two terms: the first ($\Ecal$) is a model fitting term, whereas the second is a regularization term which penalizes approximations that deviate too much from the prior. This second $\mathrm{D}_\text{KL}$ term can be computed analytically when priors and approximate posteriors have particular forms (e.g., multivariate Gaussian).

There is an apparent complication in the fact that the first term in the lower bound depends on $q(W_\eta,W_\delta,\ttheta)$ through the expectation of the log-likelihood. 
However, this is usually bypassed by employing stochastic optimization using Monte Carlo with a finite set of samples from $q(W_\eta,W_\delta,\ttheta)$:
$$
\Ecal \approx 
\frac{1}{N_{\mathrm{MC}}} \sum_{i=1}^{N_{\mathrm{MC}}} \log \left(p(Y,Z|\Xi,\psi,\Omega,W^{(i)}_\eta,W^{(i)}_\delta,\ttheta^{(i)})\right)
$$
with $W^{(i)}_\eta,W^{(i)}_\delta,\ttheta^{(i)} \sim q(W_\eta,W_\delta,\ttheta)$.
The Monte Carlo approximation is unbiased, and so it is its derivative with respect to any of the parameters of $q(W_\eta,W_\delta,\ttheta)$.
This means that we can employ stochastic gradient optimization to adapt the parameters of $q(W_\eta,W_\delta,\ttheta)$ to maximize the lower bound with guarantees to reach a local optimum of the objective \citep{Robbins51,Graves11}.
The only precaution to take to make this viable, is to reparameterize the samples from $q(W_\eta,W_\delta,\ttheta)$ using the so-called reparameterization trick \citep{Kingma14}; for instance, assuming a fully factorized Gaussian posterior over all parameters, the expression 
$
\theta^{(i)}_{\ell} = \mu_{\ell} + \epsilon_{\ell}^{(i)} \sigma_{\ell}
$ 
separates out the stochastic ($\epsilon_\ell^{(i)}\sim \Ncal(0,1)$) and deterministic ($\mu_{\ell}$ and $\sigma_{\ell}$) components in the way samples from the approximate posterior are generated.
The same can be done for the other parameters of interest $W_\eta$ and $W_\delta$.
In this way, 
the derivative with respect to the variational parameters (e.g. $\mu_{\ell}$ and $ \sigma_{\ell}$) can be used for stochastic optimization.

\paragraph{Mini-batch-based learning and automatic differentiation:}
Part of the huge success of deep learning is due to the possibility to exploit mini-batch-based learning and automatic differentiation. 
The former allows to achieve scalability, as the model progressively learns by iteratively looking at subsets of data. 
The latter allows to tremendously simplify the implementation of complex models, as one has to implement the objective function of interest and automatic differentiation takes care of computing its derivatives based on the graph of computation and the chain rule. 
Traditional implementations and approximations of \gps do not allow for the use of mini-batch learning, because doing so ignores the covariance among observations, which is crucial for effective \gp modeling. 

The proposed \gp and \dgp approximation and \svi allow us to exploit mini-batch-based learning.
When the likelihood factorizes across observations, the terms within the Monte Carlo approximation of $\mathcal{E}$, say $\mathcal{E}^{(i)}$, can be unbiasedly estimated by selecting $m$ out of $n$ terms in the set of indices $\mathcal{I}_m$ \citep{Graves11}. 
$$ 
\tilde{\Ecal}^{(i)} \approx \frac{n}{m} \sum_{j \in \mathcal{I}_m }\log \left(p(\yy_j,\zz_j |\Xi,\psi,\Omega,W^{(i)}_\eta,W^{(i)}_\delta,\ttheta^{(i)})\right)
$$
This approximation introduces an extra level of stochasticity in the optimization, but it allows one to scale the learning of these models to virtually any number of observations; previous work has reported results on \dgps for $10^7$ observations with a single-machine implementation \citep{Cutajar17}.

\paragraph{Implementation Details}
%% \noteMF{Maybe we can move this to the appendix if we need space}
Considering the large number of parameters to optimize, the learning procedure is divided in stages. 
We first focus on the computer model response: all parameters are fixed except the ones influencing the prediction of $Z$, i.e. $\sigma_z$, the means and variances of the components of $W_\eta$, the correlation length and variance of $\eta$.
%% When the likelihoods $p(z_i|\eta^*_i)$ are assumed Gaussian, the term $\Ecal$ can be derived. 
%% This was implemented to improve the optimization.
In the second stage all others parameters are freed for inferring $Y$ and $\ttheta$ jointly.
Within each stage, we first optimize the means and variances of $W$, and then all parameters jointly with a smaller learning rate.
The variational distributions are initialized equal to the priors. % 

\section{EXPERIMENTS}

In this section we extensively validate \vcal. 
For each experiment the discrepancy structure (additive~\eqref{eq:model} or general~\eqref{eq:generalModel}) will be specified.
We will also test the model when using \dgps instead of \gps. 

The experiments have the following setup. % holds for all variational model in the following experiments and are listed here:
The likelihoods $p(y_i|f_i)$ and $p(z_j|\eta^*_j)$ are assumed Gaussian with variances $\sigma^2_y$ and $\sigma^2_z$ treated as hyperparmeters within $\psi$.
All covariance functions of $\eta$ and $\delta$ are Gaussian. % \noteMF{Unify Gaussian and squared-exponential}.
The variational posteriors $q(W_\eta)q(W_\delta) q(\ttheta)$ and the prior $p(\ttheta)$ are Gaussian.

\subsection{Illustrative Example}

\begin{figure}[t]
\centering
\vspace{.0in}
\includegraphics[width=.6\linewidth,clip=true,trim=7 10 10 1]{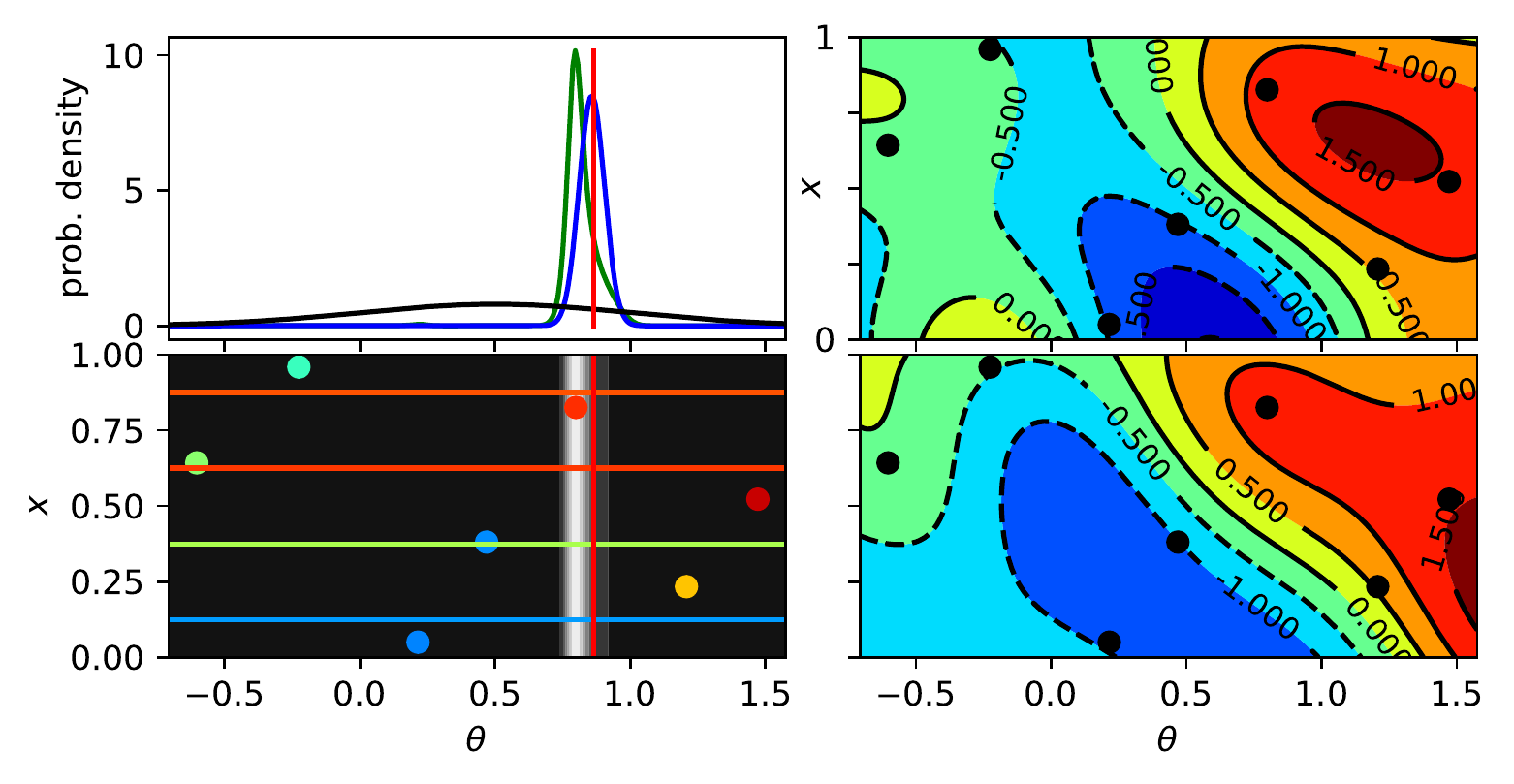}
\vspace{.0in}
\caption{{\bf Top-left:} the prior (black), the analytical posterior (green) and the variational posterior (blue) distributions of $\theta$ and
 the actual value used to generate $\YY$ (red).
{\bf Top-right:} the true response $\eta$ used to generate the data set of this example and the locations of the computer runs (dots).
{\bf Bottom-left:} shows $Y$ as horizontal lines in $\Dcal_1\times \Dcal_2$. The color of the lines correspond to the values $Y_i$. The dots represent the computer runs $Z$. The grey level represent the posterior distribution of $\theta$ (also displayed in the top-left panel).
{\bf Bottom-right:} the posterior mean of $\eta$.
}
\label{fig:variationalCalibration}
\end{figure}

We illustrate the variational calibration with one variable and one calibration input. 
As a first test, the prior and hyperparameters used to generate the data set are assumed to be known, with $\theta\sim\Ncal(0,1)$, $\sigma_\eta = 1$, $A_\eta=\frac{1}{2} I$, $\sigma_\delta = \frac{2}{10}$, $A_\delta=\frac{1}{20}$.
We choose locations for  $N=7$  computer runs and $n=4$ observations from the real process in a space filling manner in $[0,1]\times[-\frac{5}{2},\frac{5}{2}]$. 
The output vector $\ZZ$ of the computer model at $(\xx_i^*)_{i=1,\ldots,N}$ is sampled form its prior distribution. % 
In order to determine the real observations $\YY$, we first sample $p(\theta)$ to get $\theta_\text{true}$. 
Then the observation values are computed using Equation~\ref{eq:model}.
The results of \vcal are displayed in Figure~\ref{fig:variationalCalibration}. 
In the first and the third panels, we see that the posterior of $\theta$ obtained analytically by integrating out $W_\eta$ and $W_{\delta}$ has its mass concentrated around the true value $\approx0.8$, where there is a (color) match between $\ZZ$ (the the dots) and $\YY$ (the lines).
The variational posterior (blue line) offers a reasonable approximation of the true posterior. % 

\subsection{Model Calibration in Cell Biology}
We apply \vcal to a biological application, which has been previously studied in 
\citet{plumlee2017bayesian} and \citet{xie2018Bayesian}. 
The output is the normalized current through ion channels of cardiac cells needed to maintain the membrane potential at $-35$ mV. % 
The input variable $x$ is the logarithm of the experiment time rescaled to $\Dcal_1=[0,1]$. 
The calibration inputs $\ttheta\in \Dcal_2=[0,10]^3$ control a mathematical model $\eta_\text{cell}(\cdot,\ttheta)$ of the phenomenon proposed by \citet{clancy1999linking}. 
Here it is considered to be an expensive black box with $N=300$ runs available, whereas the number of observations is $n=19$. 
The runs are located in a space filling manner in $\Dcal_1\times\Dcal_2$ (Latin hypercube sampling optimized with maximin distance criterion).%

We compare \vcal with additive and general discrepancy against four competitors. 
The method ``$L_2$'' is a simple minimization over $\ttheta$ of the $L_2$ residual error $ ||\YY-\hat{\eta}_\text{cell}(\XX,\ttheta)||$, where $\hat{\eta}$ is a surrogate model of $\eta_\text{cell}$ given $X^*$, $T$ and $Z$. Its minimization takes $30$ seconds and the residual error is $1.31$.
This method is generally good for predicting observations from the real process, but it provides no quantification of uncertainty. 
%% This is a drawback in cases where the values of $\ttheta$ are important or where there is a need to detect a lack of identifiability. 
The method \koh and \name{Projected} refer to the implementation in \texttt{R} language of the \koh model
  and the Bayesian Projected Calibration of \citet{xie2018Bayesian}. 
Finally, \name{Robust}  refers to the  calibration of the \texttt{R} package \texttt{RobustCalibration} using scaled \gps \citep{gu2018Scaled}.

In Table~\ref{tab:testCase1}, we report for each method the mean squared error (\mse), $\esp_{q(\ttheta)}(||\YY-\eta_\text{cell}(\XX,\ttheta)||^2)$, where $q$ represents the estimated posterior density of $\ttheta$. 
All tuning parameters of the codes are left to default values, and all methods are run on the same machine to ensure some fairness in reporting running time (laptop with $4 \times  2.50$ GHz cores).
We see that the \mse values obtained by the methods \name{Projected}, \vcal and \name{Robust} are significantly lower than the \mse of \koh. 
The proposed \vcal is the fastest among the competitors.

\begin{table}
\caption{Comparison of calibration error on the Cell Biology calibration problem.} \label{tab:testCase1}
\begin{center}
\begin{tabular}{lll}
\textbf{METHOD}  &\textbf{TIME}&\textbf{MSE} \\
\hline \\
{\name{Projected}}             & 3792 &2.52 \\
\vcal additive            & 79  &1.83\\
\vcal general            & 93  &1.55\\
\koh           & 3245  &5.10\\
{\name{Robust} }           & 361&1.99\\
\end{tabular}
\end{center}
\end{table}

\begin{figure}[t]
\vspace{.0in}
\begin{center}
\includegraphics[width=.4\linewidth,clip=true,trim=0 0 0 2]{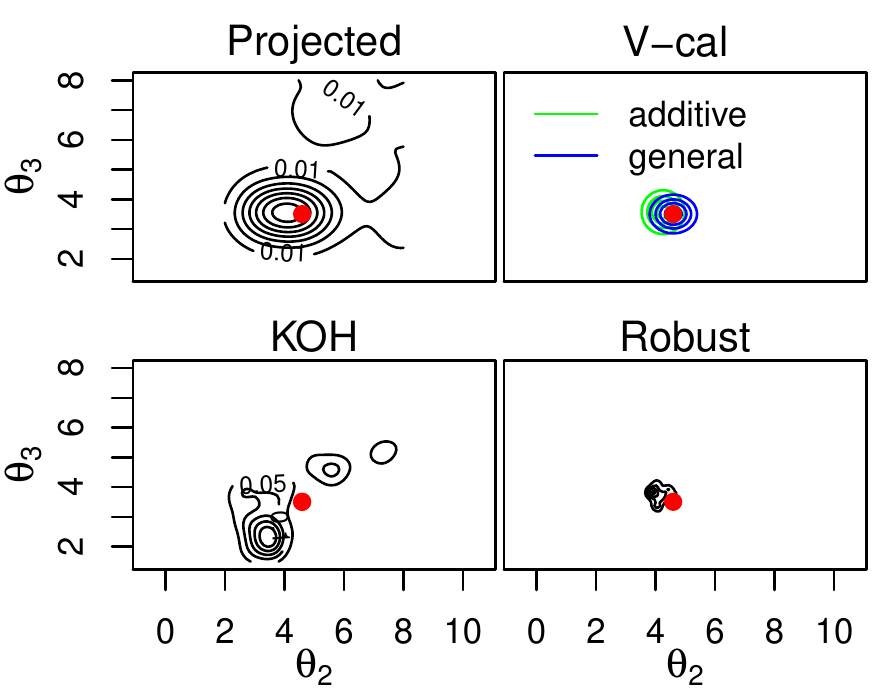}
\end{center}
\vspace{.0in}
\caption{Posterior distributions of $\ttheta$ from the calibration methods (integrated over $\theta_1$ for visualization)}
\label{fig:testCase2Esti}
\end{figure}

\begin{figure}[t]
\vspace{.0in}
\begin{center}
\includegraphics[width=.65\linewidth,clip=true,trim=0 0 0 12]{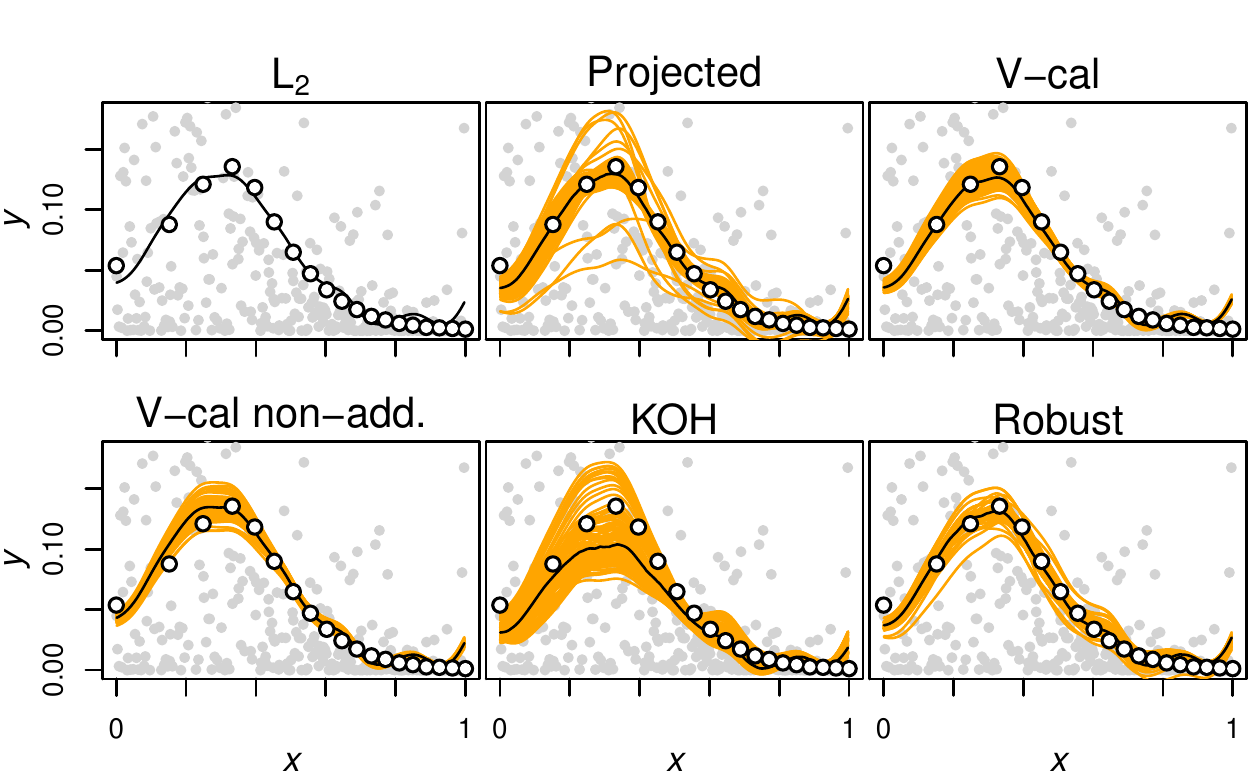}
\end{center}
\vspace{.0in}
\caption{Samples of $\eta_\text{cell}(\cdot,\ttheta)$, with $\ttheta$ drawn from its  posterior. The grey dots represent the computer runs and the white dots the real observations.}
\label{fig:testCase2Pred}
\end{figure}

\begin{figure}[t]
\vspace{.0in}
\begin{center}
\includegraphics[width=.6\linewidth,clip=true,trim=1 1 1 0]{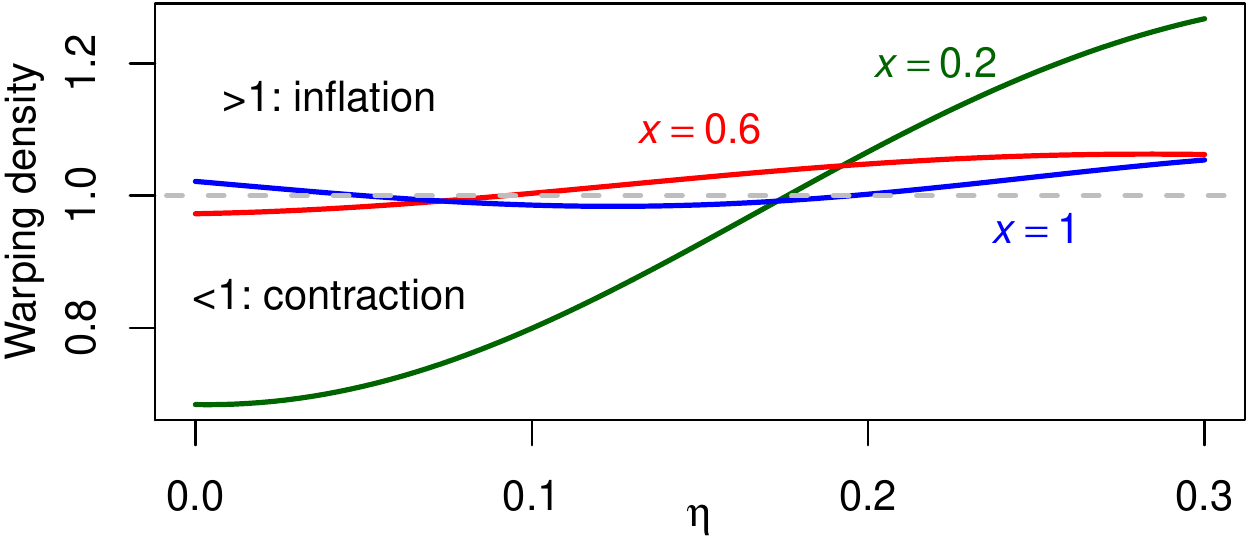}
\end{center}
\vspace{.0in}
\caption{Derivative $\frac{\partial g(\cdot,x)}{\partial \eta}$ for three values of $x$
}
\label{fig:warping}
\end{figure}

The posterior distributions over $\ttheta$ obtained by the calibration methods we tested are reported in Figure~\ref{fig:testCase2Esti}. 
All methods yield a distribution concentrated around the $L_2$ minimizer (the red dot).
However, the distributions are clearly not similar to each other (except for \name{Robust} and \vcal). 
This could be explained by differences in the model formulations. % 
Although we ensured that the covariance and mean functions are the same for all competing methods (Mat\'ern with smoothness 5/2 and constant mean), there are several differences that cannot be matched. % across methods. 
For instance, \name{Robust} has an additional step in the hierarchy of priors concerning the $L_2$ norm of the discrepancy. 
Moreover, the definition of the calibration parameters $\ttheta$ itself differs among methods. 
In \name{Projected},  $\ttheta$ is a minimizer of a given stochastic process, while other methods follow the \koh definition. 
Also \name{Robust} performs a fully Bayesian inference including hyperparameters, while in the other methods, including ours, they are optimized.
It would be straightforward to allow for a Bayesian treatment of the hyperparameters in \vcal, but we leave this for future work. 
To visualize the results of the calibration process, in Figure~\ref{fig:testCase2Pred} we overlay the observations from the real process with the responses of the computer model $\eta_\text{cell}(\cdot,\ttheta)$ when $\ttheta$ is sampled from its posterior distribution. 
All the probabilistic method present a good fit while allowing for quantification of uncertainty in predictions, with larger uncertainty for models that account for the uncertainty in the hyperparameters. 
We see how the computer model output $\eta$  is warped by $g$ in $\vcal$ with general discrepancy (Equation~\eqref{eq:generalModel}). In Figure~\ref{fig:warping} we display the expected derivative of the warping with respect to the computer model output, i.e.  $\esp\frac{\partial g(\cdot,x)}{\partial \eta}$, for three values of $x$.  As the estimated values oscillate around one for every $x\in\Dcal_1$, this model confirms that an additive discrepancy is a sensible assumption. When the estimated  $g(\cdot,x)$ is exactly the identity, the general discrepancy boils down to an additive one.  This figure also shows how the model with general discrepancy can adapt to data sets with space dependent behavior. Indeed in this test case the values of $\eta$ have a very different distribution according to $x$. If $x$ is around 0.2, the distribution of the computer runs is very asymmetric, with a heavy tail for high values, and very short for low value (see the grey dots in Figure~\ref{fig:testCase2Pred}). On the other hand for higher values of $x$, say higher than 0.6, the distribution of $Z$ is more symmetric, and looks closer to a Gaussian distribution. This corresponds to the warping observed in Figure~\ref{fig:warping}, where $\eta$ gets its output warped and concentrated asymmetrically toward lower values for $x=0.2$, while for $x=0.6$ or $1$, its Gaussian output is almost left untouched.

\subsection{Model Calibration for Complex Response}
We deal now with a case-study with locally non-smooth response of the computer model, for which a stationary \gp is generally inadequate.
The computer model is a simulator of the effects of underground nuclear tests on radionuclide diffusion into aquifers at the Yucca Flats in the United States \citep{fenelon2005analysis}. 
We take the same data set as generated by a script in the supplementary material of \citet{pratola2016Bayesian}, which is available online, with $d_1 = 2$, $d_2=6$, $n=10$ and $N=17600$.

In \citet{pratola2016Bayesian}, the size of the dataset as well as nonstationary modeling is handled with a sum-of-trees regression. 
We carry out calibration using \vcal with a two-layer \dgp emulator for the computer model to showcase the ability of a more complex emulator to capture the nonstationarity that characterizes this problem.
We therefore compare \vcal with a shallow \gp emulator.  The implementation details on the initialization can be find in the supplement.
Furthermore, we compare against the modularized method with Local Approximate \gps (\lagp) of \citet{gramacy2015Calibrating}.

\begin{figure}[t]
\vspace{.0in}
\begin{center}
\includegraphics[width=.6\linewidth,trim=1 1 1 0]{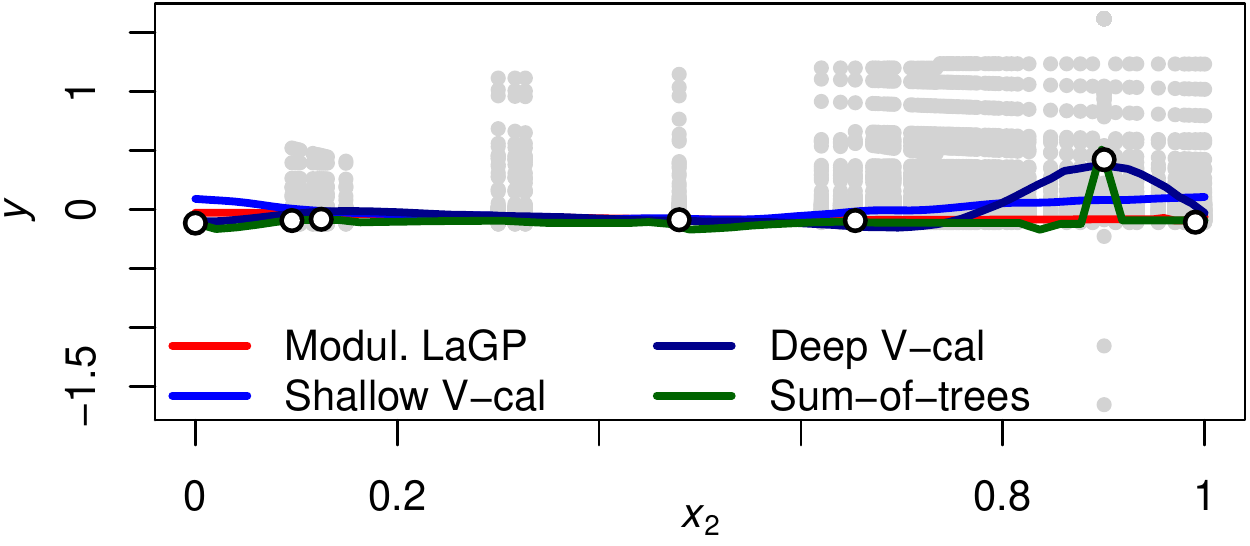}
\end{center}
\vspace{.0in}
\caption{Mean of the posterior over the function $f(\cdot,\ttheta)$ modeling the real observations}
\label{fig:nevada}
\end{figure}

In Figure~\ref{fig:nevada}, we display the posterior over the function $f(\cdot,\ttheta)$ modeling the real observations. 
We observe that only the deep variational calibration method and the sum-of-trees approach manage to reproduce the nonstationary nature of the data set by capturing the ``spike'' characterizing one of the observations.

\subsection{Data Set Size Scalability}

We now showcase the scalability of \vcal to a large calibration problem in $8$ dimensions with one million computer runs, and $100,000$ real observations. 
We use the borehole function $\eta_\text{bh}(\xx,\tt)$,  $\xx\in[0,1]^5$ $\tt\in[0,1]^3$, which is a widely used function in the literature of computer experiments. 
The discrepancy function is a rational function (as e.g. in \citet{gramacy2015Calibrating})  (see supplement for details).
We are interested in retrieving a randomly chosen true value $\ttheta = [0.089,0.308,0.372]\trans$. 
The locations $X$, $X^*$ and $T$ are generated with Latin hypercube sampling.
To generate $\YY$, a white Gaussian noise $\vvarepsilon$ of standard deviation $\sigma_\text{bh} = 5\times10^{-3}$ is added: $y_i = \eta(\xx_i,\ttheta)+\delta(\xx_i)+\varepsilon_i$.

We build a shallow \vcal model with additive discrepancy as described by Equations~\eqref{eq:model}\eqref{eq:rfEta}\eqref{eq:rfDelta}. 
The covariance functions for the centered \gps $\eta$ and $\delta$ are isotropic Gaussian approximated by $100$ random features. 

Concerning the sum-of-trees calibration, a  sensible computation budget would be to set $2000$ posterior samples plus $10000$ for burn-in, with $1000$ tree cutpoints. 
However, this corresponds to one month of computation on our computers, so we divided the sampling budget by $5$, and set $100$ cutpoints, keeping all other default parameters untouched.

We did not compare with the modularized calibration using \lagp, as the current implementation in \texttt{R} does not support large amount of real observations. 
This does not question the relevance of the method, which could be fixed by using a scalable \gp for the discrepancy.

\begin{figure}[t]
\vspace{.0in}
\begin{center}
\includegraphics[width=.6\linewidth]{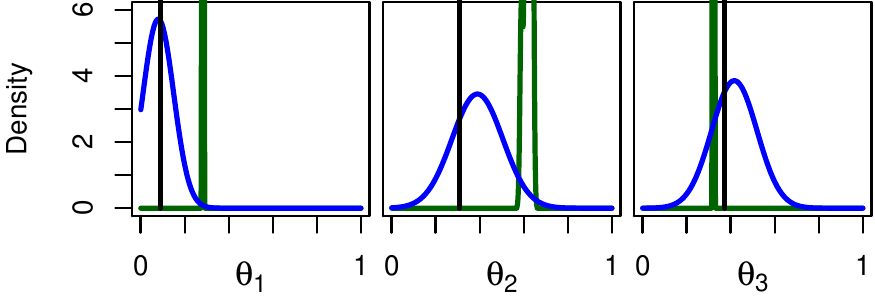}
\end{center}
\vspace{.0in}
\caption{Posterior distribution of $\ttheta$ on a large data set (black: truth, blue: \vcal, green: sum-of-trees).} 
\label{fig:borehole}
\end{figure}

We evaluate the performance by comparing the posterior of $\ttheta$ with the truth (Figure~\ref{fig:borehole}) and by evaluating the \mse error between the computer model and observation from the real process $\esp_{q(\ttheta)}(||\YY-\eta_\text{bh}(\XX,\ttheta)||^2)$ (Table~\ref{borehole-table}).
\vcal provides the best performance both in retrieving $\ttheta$ and \mse, and it is the fastest by far. 

\begin{table}
\caption{Results of calibration on a large data set} \label{borehole-table}
\begin{center}
{\begin{tabular}{lcc}
   \textbf{CALIBRATION }    & \textbf{TIME}~(h) & \textbf{MSE}  \\
   \hline \\
None (unif. samples on $\Dcal_2$) & 0&0.32\\
\vcal & 1.4 & 0.03\\
Sum-of-trees        & 132.4 &   0.14 \\ 
\end{tabular} 
}
\end{center}
\end{table}

\section{CONCLUSIONS}

The \koh model and inference in \citet{kennedy2001Bayesian} offers a classical framework to tackle calibration problems where quantification of uncertainty is of primary interest. 
In this paper we proposed a number of improvements over the \koh calibration model and inference.

From the modeling perspective, we cast the \koh calibration model as a special case of a more general \dgp model, where the latent process modeling the real observations is a warped version of the emulator of the computer model. 
In the experiments, we show that this general calibration model retains the possibility to reason about uncertainty in the mismatch between the computer model and the real process. 
Furthermore, the proposed approximation of \gps and \dgps with random features and inference through variational techniques gives us a number of advantages, namely the possibility to extend to the use of \dgps to emulate complex computer model outputs, and scalability as demonstrated in the experiments. 
Crucially, this yields an extremely practical calibration approach that can be implemented using modern development frameworks featuring automatic differentiation. 

We are currently investigating the issue of non-identifiability in the context of \vcal.
We are also extending \vcal to handle cases where the uncertainty in the model can be used to guide the incremental design of the experiment. 
Finally, we are investigating the application of \vcal to other large-scale calibration problems in environmental sciences, where the \koh model and related calibration methodologies are usually not the preferred choice due to its limited scalability. 

\section*{Acknowledgment}
Maurizio Filippone gratefully acknowledges support from the AXA Research Fund.

\bibliographystyle{abbrvnat_nourl}
\bibliography{biblio}

\newpage
\appendix

\section*{Appendix}

\section{Borehole objective and  discrepancy functions}
The objective function used in section 4.4 is defined for all  $\xx\in[0,1]^5$ and $\tt\in[0,1]^3$, with
\begin{align}
\eta_\text{bh}(\xx,\tt) &= \frac{2 \pi T_u (H_u-H_l)}{\ln(r/r_w)\left(1+\frac{2LT_u}{\ln(r/r_w)r_w^2K_w}+\frac{T_u}{T_l}\right)}\nonumber,\\
\delta_\text{bh}(\xx) &= \frac{2  (10  x_1^2 + 4  x_2^2)}{50  x_1  x_2 + 10}\nonumber,
\end{align}
with $T_u = x_1(115600 - 63070) + 63070$, $H_u = x_2(1110 - 990) + 990$, $H_l=x_3(820 - 700) + 700$, $L=x_4(1680 - 1120) + 1120$, $K_w=x_5(12045 - 9855) + 9855$, $r_w=t_1(0.15 - 0.05) + 0.05$, $r=t_2(50000 - 100) + 100$, $T_l =t_3(116 - 63.1) + 63.1$.

%\section{Computer Runs of the Cell Biology Test Case}
%In section 4.2, we illustrate the warping of  model output $\eta$ of the Cell Biology test case. To underline how the 
%
%We see how the computer model output $\eta$  is warped by $g$ in $\vcal$ with general discrepancy (Equation~\eqref{eq:generalModel}). We display in Figure~\ref{fig:warping} the expected derivative of the non-additive discrepancy with respect to the computer model output, i.e.  $\esp\frac{\partial g(\cdot,x)}{\partial \eta}$, for three values of $x$.  As the estimated values oscillate around one for every $x\in\Dcal_1$, this model confirms that an additive discrepancy is a relevant assumption. Indeed, when  the estimated  $g(\cdot,x)$ is exactly the identity, the general discrepancy boils down to an additive one.  This figure also shows how the model with general discrepancy can adapt to data sets with space dependent behavior. Indeed in this test case the values of $\eta$ have a very different distribution according to $x$.
%
%\begin{figure}[h]
%\vspace{.0in}
%\begin{center}
%\includegraphics[width=.4\linewidth,clip=true,trim=0 0 0 0]{{"../code/Comparisons/test case 2/plots/Z"}.pdf}
%\end{center}
%\vspace{.0in}
%\caption{Computer Runs of the Cell Biology Test Case}
%\label{fig:Z}
%\end{figure}

\section{Initialization of variational parameters in \vcal for the Radionuclide model}

We list in the following table the settings of the $\vcal$ models used for the experiments of the section 4.3.
\begin{table}[h]
\caption{Radionuclide Model: Initial Values for the \vcal models} \label{tab:testCase1}
\begin{center}
\begin{tabular}{lcc}
\textbf{PARAM.}  &\textbf{SHALLOW} &\textbf{DEEP} \\
\hline \\
$\esp_{q(\ttheta)}(\ttheta)$        & $\frac{1}{2}[1,1,1]\trans$ & $\frac{1}{2}[1,1,1]\trans$\\
$\var_{q(\ttheta)}(\ttheta)$      & $\frac{1}{4}[1,1,1]\trans$ &$\frac{1}{4}[1,1,1]\trans$ \\
$\sigma_y$             &  $10^{-2}$ &$10^{-2}$\\
$\sigma_z$     &   $10^{-3}$ &  $10^{-3}$\\
$A_\eta$, $A_\delta$     &     $20 I$ &$20 I$\\
$\sigma_\eta$, $\sigma_\delta$     &   $1$, $\frac{1}{10}$ &   $1$, $\frac{1}{10}$\\
$A_{\text{layer}~i}$    &  $\emptyset $& $2 I_{d_1+d_2}$\\
$\sigma_{\text{layer}~i}$    &  $\emptyset $& $1$\\
\end{tabular}
\end{center}
\end{table}

\end{document}